\def\year{2020}\relax
\documentclass[letterpaper]{article} 
\usepackage{aaai20}  
\usepackage{times}  
\usepackage{helvet} 
\usepackage{courier}  
\usepackage[hyphens]{url}  
\usepackage{graphicx} 
\usepackage{graphicx}  
\usepackage{amsmath}
\usepackage{amsfonts}
\usepackage[noend]{algpseudocode}
\usepackage{algorithmicx,algorithm}
\usepackage{booktabs}
\usepackage{multirow}

\title{Generating Persona Consistent Dialogues by Exploiting Natural Language Inference}
\author{\Large \textbf{Haoyu Song, Wei-Nan Zhang, Jingwen Hu, Ting Liu}\\ 
Research Center for Social Computing and Information Retrieval\\ 
Harbin Institute of Technology, Heilongjiang Province, China\\
\{ hysong, wnzhang, jwhu, tliu \}@ir.hit.edu.cn 
}

\begin{document}

\maketitle

\begin{abstract}
  Consistency is one of the major challenges faced by dialogue agents. A human-like dialogue agent should not only respond naturally, but also maintain a consistent persona. In this paper, we exploit the advantages of natural language inference (NLI) technique to address the issue of generating persona consistent dialogues. Different from existing work that re-ranks the retrieved responses through an NLI model, we cast the task as a reinforcement learning problem and propose to exploit the NLI signals from response-persona pairs as rewards for the process of dialogue generation. Specifically, our generator employs an attention-based encoder-decoder to generate persona-based responses. Our evaluator consists of two components: an adversarially trained naturalness module and an NLI based consistency module. Moreover, we use another well-performed NLI model in the evaluation of persona-consistency. Experimental results on both human and automatic metrics, including the model-based consistency evaluation, demonstrate that the proposed approach outperforms strong generative baselines, especially in the persona-consistency of generated responses. Our codes are available at: https://github.com/songhaoyu/RCDG.
\end{abstract}

\section{Introduction}
\begin{figure}
    \centering
    \includegraphics[width=1.0\linewidth]{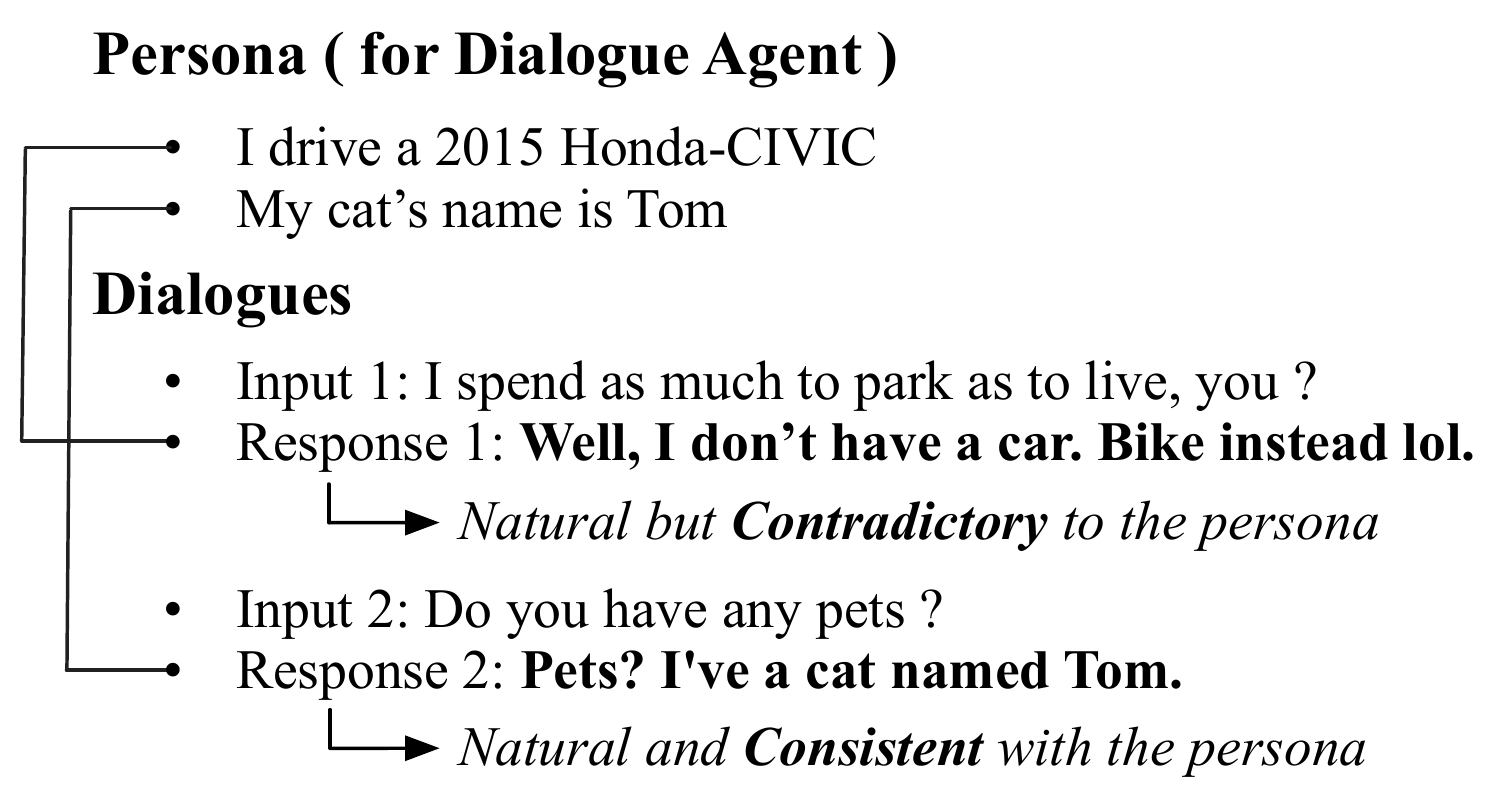}
    \caption{Naturalness is an important attribute of dialogue responses. In persona-based dialogue generation, the consistency with persona is another essential factor to consider. An ideal response should be not only natural but also consistent with the persona.}
    \label{fig:persona_and_dialogues}
  \end{figure}

  Despite the recent success of dialogue generation in open-domain by training from large volumes of human-to-human interaction data~\cite{shang2015neural,serban2016building,li2017adversarial,zhu-etal-2019-retrieval}, conversing to a dialogue agent is still in its infancy, and one major issue for these data-driven models is the lack of a consistent persona~\cite{vinyals2015neural,li2016persona,zhang2018personalizing,ijcai2019-721}. Figure~\ref{fig:persona_and_dialogues} shows how consistency affects the quality of dialogues.

  One practical approach to increase the consistency of a dialogue agent is to explicitly define a set of personal facts describing the characters (the {\it personas}) of the agent and learn to generate responses that reflect the predefined personas~\cite{zhang2018personalizing}. However, due to the lack of consistency modeling and the maximum-likelihood estimation (MLE) objective function, these persona-based models still face the inconsistency issue~\cite{WelleckDNLI}.

  Natural Language Inference (NLI) learns a mapping between a sentence pair and an entailment category. Taking advantages of the NLI techniques in natural language understanding~\cite{bowman2015large}, the detection of persona-consistency can be modeled as an NLI task~\cite{WelleckDNLI}, which assigns a label of  {\it entailment}, {\it neutral}, or {\it contradiction} to an ``(utterance, persona)'' pair. 
  Meanwhile, existing persona-based dialogue models are limited by their loss functions. For these deep generative models, it is difficult to design a differentiable training objective to
  exploit the NLI based consistency detection method.
  Besides designing a differentiable training objective, reinforcement learning (RL) offers another solution to this problem, which backpropagates the reward signals to guide the generator.

  In this paper, different from re-ranking the archived responses~\cite{WelleckDNLI}, we take advantages of the NLI techniques in guiding the generation of persona consistent dialogues. 
  Specifically, we propose a system trained using reinforcement learning. Our model has one evaluator with two modules and one generator.
  The evaluator consists of a naturalness module and a consistency module. The naturalness module is trained adversarially for higher accuracy. As for the consistency module, we use an NLI styled classifier to detect the consistency between responses and personas. We further employ two different NLI classifiers in our experiments to investigate the role of NLI signals.
  The generator, which is a persona-based attentive Seq2Seq model~\cite{zhang2018personalizing}, takes message and persona texts as input and generates responses that reflect the persona texts. Note that more advanced generative models such as MASS~\cite{MASS} can also be exploited as our generator.


  We summarize the contributions as follows:
  \begin{itemize}
  \item We propose an RL framework for persona consistent dialogue generation, thus addressing the challenge of training objective need to be differentiable in persona-based dialogue models.
  \item To the best of our knowledge, this is the first work that exploits NLI techniques to enhance the generation of persona consistent dialogues.
  \item Evaluations are carried out both quantitatively and qualitatively, and experimental results show that our model outperforms strong baselines, especially in terms of persona-consistency.
  \end{itemize}

\section{Related Work}
\subsubsection{Persona-based Dialogue Generation}
In open-domain dialogue generation, \citeauthor{zhang2018personalizing} \shortcite{zhang2018personalizing} initiate a new line of research (the persona-based dialogue) by introducing the Persona-Chat dataset, with explicit persona information in each dialogue session. They further propose two generative models, {\it persona-Seq2Seq} and {\it Generative Profile Memory Network}, to incorporate persona texts into responses. In the persona-based scenario, a model is associated with a persona, which is composed of several persona texts (See the top two sentences in Figure~\ref{fig:persona_and_dialogues}). A response is then generated using both the input message and the assigned persona. Following this line, \citeauthor{deepcopy} \shortcite{deepcopy} apply the {\it DeepCopy} model in the persona-based dialogue generation. These works have laid a solid foundation for this area. Through attention or copy, generated responses can reflect the predefined persona. However, the loss functions in these models do not take the consistency issue into account, and inconsistency is still a problem to be addressed in the existing approaches \cite{WelleckDNLI}.

\subsubsection{Natural Language Inference}
The task of Natural Language Inference (NLI) is to learn a function~$f_{NLI}(p,h)\rightarrow\{\text{E},\text{N},\text{C}\}$, where $p$ and $h$ denote {\it premise} and {\it hypothesis} respectively.
The output $\text{E}$, $\text{N}$ and $\text{C}$ represent {\it entailment}, {\it neutral} and {\it contradiction} between the {\it premise} and {\it hypothesis}. 
Since the release of large scale corpus SNLI~\cite{bowman2015large}, deep neural network methods have made promising progress~\cite{chen2017ESIM,gong2018DIIN,kim2019DRCN}. \citeauthor{WelleckDNLI} \shortcite{WelleckDNLI} model the detection of dialogue consistency as an NLI task and propose the DNLI (Dialogue NLI) dataset, which is similar to the SNLI but in the domain of persona-based dialogue. Further, they verify the effectiveness of using NLI model to re-rank candidate responses in a retrieval-based dialogue model. Compared with the retrieval-based model, the responses from generative models are not limited to the given dataset. Moreover, exploiting consistency detection method in deep generative dialogue models has not been explored yet.

\subsubsection{Reinforcement Learning} In recent years, deep reinforcement learning has been widely applied in natural language processing, such as machine translation \cite{wu-etal-2018-study}, visual question generation \cite{fan2018reinforcement}, paraphrase generation \cite{li2018paraphrase}, anaphora resolution \cite{yin2018deep} etc. The advantage of reinforcement learning lies in that it does not need a differentiable objective function. In open-domain dialogue generation, \citeauthor{li_deeprl} \shortcite{li_deeprl} manually defined three rewards and use reinforcement learning to train the dialogue agent. Further, \citeauthor{li2017adversarial} \shortcite{li2017adversarial} apply adversarial learning method \cite{yu2017seqgan} for dialogue generation and propose the {\it REGS} model. This model shows its strength in the naturalness of generated responses. However, natural responses can also be inconsistent, especially in the persona-based scenario (as shown in Figure~\ref{fig:persona_and_dialogues}).

\begin{figure*}[ht]
  \centering
  \includegraphics[width=1.0\linewidth]{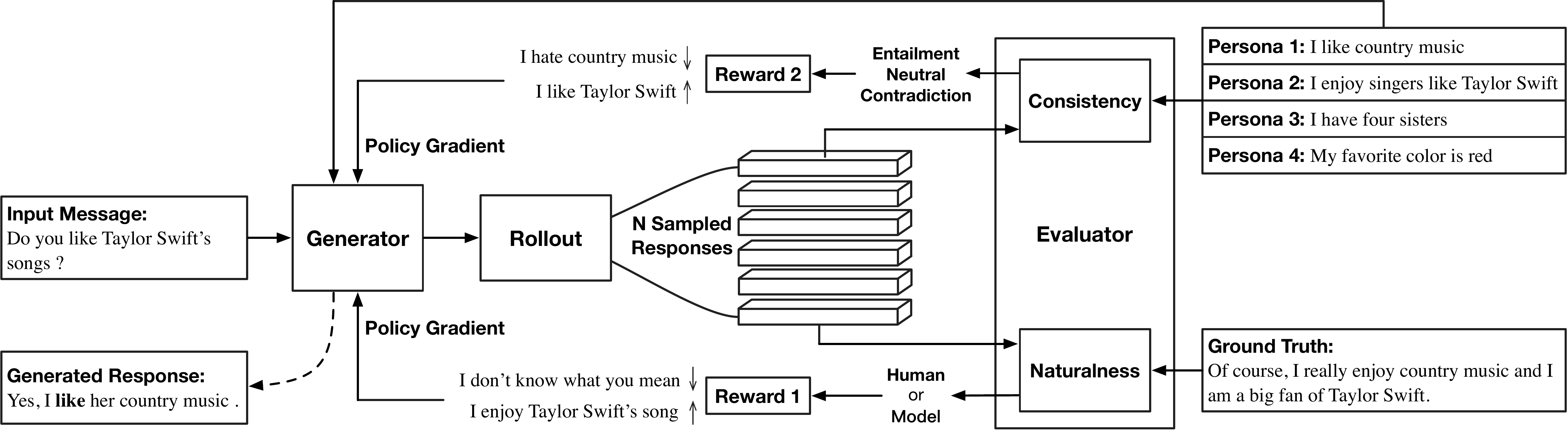}
  \caption{The overall framework of our model, which mainly consists of a generator and an evaluator. The dashed connection only appears in the generation process. The $\uparrow$ and $\downarrow$ denote that the generation of a response is encouraged and discouraged by the reward signals respectively.}
  \label{fig:model_main}
\end{figure*}

\section{Proposed Approach}
\label{sec_PA}

\subsection{Problem Definition}
\label{sec_TaskDef}

Our goal is to learn a generative model $G$ to deliver persona consistent dialogues, which can be formally defined as: given an input message $X$, and a set of persona texts $P={\{P_1,P_2,...,P_n\}}$, to generate a response $\hat Y$, based on both input message $X$ and the persona text set $P$, i.e., $\mathcal{G}(X,P)={\hat Y}$. Moreover,  $\hat Y$ should be consistent with the persona text set $P$, which means the NLI category between $\hat Y$ and any $P_i$ should be {\it entailment} or {\it neutral}, rather than {\it contradiction}, i.e., $\forall P_i \in P$, $\mathit{NLI}({\hat Y}, P_i)\in\{\text{E}, \text{N}\}$, where $\text{E}$ and $\text{N}$ denote $entailment$ and $neutral$ respectively.

\subsection{Evaluator}
\label{sec_Bi_dis}
The proposed reinforcement learning framework consists of two components: an evaluator and a sequence generator, as illustrated in Figure~\ref{fig:model_main}.

As forementioned, an ideal response should be not only natural but also consistent with personas. Therefore, we consider these two attributes of responses while training the generator. More concretely, whether a response is as natural as from human ({\it natural} or {\it unnatural}) and whether a response is consistent with predefined personas ({\it entailment}, {\it neutral} or {\it contradiction}). These two attributes are independent of each other, and an ideal response $Y^*$ should satisfy:
\begin{align}
\label{ideal_Y}
  Y^*\in {Natural} \cap {Entailment}.
\end{align}
The key idea is to encourage the generator to generate responses that satisfy Formula~(\ref{ideal_Y}). We use the policy gradient method in reinforcement learning to train the generator. We will discuss this in detail later.

Notice that our evaluator consists of two modules, rather than one jointly trained module, which is due to the difference between the two attributes. For the naturalness module, it can benefit from the adversarial training scheme \cite{yu2017seqgan,li2017adversarial}, as naturality is reflected in the training data. Naturalness as a submodule in the evaluator can achieve higher accuracy with adversarial training. In contrast, no labels are available in the training process to improve the performance of natural language inference.

\subsubsection{{Naturalness Module}}
The naturalness module $E_{N}$ is proposed to distinguish between human responses and model generated ones. 
As the generator is updating during the training process, new examples from models are generated. Therefore, we shall update the $E_N$.

It is safe to assume that responses from humans are always more natural than the ones from models. From this observation,  we take responses from the training data as positive examples and responses from the generator $G$ as negative examples. $E_N$ guides the sequence generator $G$ to predict responses closer to the examples from the training data, which is more natural.

In more detail, the naturalness module $E_N$ is a binary classifier that takes response $Y$ or $\hat Y$ as input\footnote{In our experiments, we found that the way of taking $\{X, Y\}$ as input to $E_N$ didn't bring significant performance improvements in the accuracy of $E_N$, so we choose the more straightforward way.} and produces a softmax distribution over two classes, indicating whether the response is from human (natural) or model (unnatural).
The input is encoded into a vector representation using a bidirectional GRU encoder, which is then fed to a highway architecture, followed by a fully-connected layer with two-class softmax function.

The objective function of $E_N$ is to minimize the cross-entropy between the ground truth label and the predicted probability.
And the reward from $E_N$ is:
\begin{align}
\label{r1}
  R_1 = E_N^{+},
\end{align}
where $E_N^+$ is the output probability of $\hat Y$ from the human.

\subsubsection{{Consistency Module}}
The consistency module $E_C$ is an NLI classifier. We introduce this module to detect the consistency in dialogues by distinguishing \{{\it entailment}, {\it neutral}, {\it controdiction}\} between generated responses and the persona texts.
Recent NLI models~\cite{conneau2017infersentence,chen2017ESIM,gong2018DIIN,kim2019DRCN} are usually trained on large-scale datasets like SNLI~\cite{bowman2015large}. The domain adaption problem could lead to a performance gap. Therefore, a better dataset for our task is the recently released DNLI~\cite{WelleckDNLI}, which is in the persona-based dialogue domain.

The consistency module $E_C$ is not updated in the adversarial training process of $E_N$. Due to the assumption that responses from humans are natural, $E_N$ can always get positive examples (the human responses from training data) and negative examples (the generated responses from $G$) during the adversarial training process. 
However, as exemplified in Figure~\ref{fig:persona_and_dialogues}, a natural response does not necessarily entail persona texts and vice versa. Due to this difference, $E_C$ cannot be iteratively updated like $E_N$.

In addition to exploiting NLI models in dialogue generation, another issue worth exploring is how the performance of different NLI signals affects the quality of dialogue generation. Thus in our experiments, we apply two NLI classifiers with performance differences:

\begin{itemize}
\item {\bf Base Model} We use the GRU to learn the sentence representations and then put them into a multilayer perceptron (MLP) classifier. The MLP has a hidden layer with {\it tanh} activation and a softmax output layer in our experiments. For training, we use a multi-class cross-entropy loss. In the following sections, we abbreviate this model as $E_{base}$.

\item {\bf Finetuned BERT} With multilayer bidirectional Transformers~\cite{vaswani2017attention}, BERT~\cite{devlin2018bert} has achieved state-of-the-art results on various natural language understanding tasks, including NLI. We finetune the BERT$_{base}$ model on the DNLI dataset and achieve best results compared with several other reported results on this dataset. In the following sections, we abbreviate this model as $E_{bert}$.
\end{itemize}

Finally, we can get the three-class confidences from the output layer of a consistency module. The reward from $E_C$ can formulate as:
\begin{align}
\label{r2}
  R_2 = max_i \text{E}_i - max_j \text{C}_j, i,j \in \{1,2,...,|P|\},
\end{align}
where $\text{E}$ is the confidence for {\it entailment} and $\text{C}$ is the confidence for {\it contradiction}. Index $i$ (or $j$) denotes the confidence is calculated between $\hat Y$ and $P_i$ (or $P_j$). This reward is designed to encourage {\it entailment} and discourage {\it contradiction} between $\hat Y$ and $P$.

\subsection{Reinforcement Learning}

We formalize the persona consistent dialogue generation problem as a reinforcement learning task. 
That is, we train a generator $G$ to produce a response $\hat Y_{1:t}=\{y_1,y_2,...,y_t\}$, where $y_i$ represents a word in the vocabulary. 
At each timestep {\it t}, the state $s_t$ is the current produced word $(y_1,y_2,...,y_{t-1})$, and the action $a$ is the next selected word $y_t$. The policy model $G(Y_t|Y_{1:t-1})$ defines the probability that selecting the $t$-th word depending on the previously generated words, which is the current state.

\subsubsection{{Sequence Generator}}
Our generator $G_{\theta}$ takes a form similar to Seq2Seq model with attention mechanism. The only difference is that we prepend persona texts to the input sequence, i.e., $X = \forall P_i \in P||X$, where $||$ denotes the concatenation. The same strategy is also applied to the generative model in \citeauthor{zhang2018personalizing} \shortcite{zhang2018personalizing}.

\subsubsection{{Reward Estimation}}
In reinforcement learning, the training objective is to maximize the accumulated future rewards.
We encourage the generator to generate responses that are close to human and consistent with the predefined persona.
Based on rewards from the naturalness module and consistency module, the final reward function is:
\begin{align}
\label{final_R}
 {\mathcal R_{E_{\phi}}}(\hat Y|X,P) = \lambda R_1 + (1-\lambda) R_2,
\end{align}
where ${E_{\phi}}$ is our evaluator with the parameter $\phi$. We train the generator $G_{\theta}(y_t|Y_{1:t-1})$ to generate a response from the initial state $s_0$ to maximize its expected final reward:
\begin{align}
\label{RF_obj}
 {\mathcal J}(\theta) 
 &= \sum\limits_{t=1}\limits^{T}G_{\theta}(\hat y_t|s_{t-1}) \cdot Q_{E_{\phi}}^{G_{\theta}}(\hat y_t,s_{t-1}),
\end{align}
where $Q_{E_{\phi}}^{G_{\theta}}(\hat y_{t},s_{t-1})$ is the action-value function at timestep $t$. When there is a finished response $\hat Y_{1:T}$, the evaluator can provide a reward by Eq.~(\ref{final_R}) for the action-value function:
\begin{align}
\label{Q_R}
Q_{E_{\phi}}^{G_{\theta}}(\hat y_T,s_{T-1}) = {\mathcal R_{E_{\phi}}}(\hat Y_{1:T}|X,P).
\end{align}

\subsubsection{{Rollout Policy}}
Our evaluator is trained to predict based on a complete sequence. Thus the reward from Eq.~(\ref{Q_R}) can only be used for the final states in a response (the generation of a response must be finished), which will hurt the effectiveness of training the generator.

To evaluate the action-value $Q$ at an intermediate state $s_t$ (t $<$ T), a common strategy is to apply rollout policy, such as Monte Carlo search, to sample the last T $-$ t words for the partially decoded response $\hat Y_{1:t}$ \cite{yu2017seqgan,li2017adversarial,fan2018reinforcement}. 
When applying a rollout policy, the model keeps sampling words from the generative model until the decoding is finished. This process is repeated for $N$ times, and the average reward of the sampled responses by Eq.~(\ref{final_R}) is used as the action-value for the state $s_t$:
\begin{align}
\label{Equ_MC_R}
&Q_{E_{\phi}}^{G_{\theta}}(\hat y_{t+1},\hat s_{t}) = \frac{1}{N}\sum\limits_{i=1}\limits^{N}{\mathcal R_{E_{\phi}}}(\mathbf{rollout}_{G_{\theta}}^{\ i}(\hat Y_{1:t+1})|X,P).
\end{align}
With rollout policy, the gradient of Eq.~(\ref{RF_obj}) can be solved by policy gradient method:
\begin{align}
\label{Equ_gradient}
\nabla_{\theta}{\mathcal J}(\theta) = \sum\limits_{t=1}\limits^{T}{\mathbb E}[\ \nabla_{\theta} log G_{\theta}(\hat y_t|s_{t-1})
\cdot\ Q_{E_{\phi}}^{G_{\theta}}(\hat y_{t},s_{t-1})\ ],
\end{align}
and the expectation ${\mathbb E}$ can be approximated by sampling methods.

When $N$ are large enough, MC search leads to a reasonable estimate of the sentence rewards. However, this comes at a high computational cost. When $N$ decreases for the balance of computational time, the diversity of the sampled responses are affected, which could lead to a poor estimate. Therefore, we propose a different rollout policy:
1. at step $t$,  the model first generates a $t$ words' subsequence with beam search.
2. at step $t+1$, the model keeps $N$ different words with the top probabilities.
3. after step $t+1$, the model continues to generate words with a sampling-based method for the partially decoded sequences with $t+1$ words.

We apply this rollout policy for a balance of the computational time and the sample diversity. In this way, we can get diverse samples, even when $N$ is relatively small.

\subsection{Adversarial Training}
\label{sec_PG}

\begin{algorithm}[t]
\label{algorithm_1}
\caption{Sketch of the training procedure}
\hspace*{0.02in} {\bf Requires:}
generator $G_{\theta}$, evaluator $E_N$ and $E_C$,\\
\hspace*{0.22in}  dialogue corpus $\mathcal S$, nli dataset $\mathcal L$.

\begin{algorithmic}[1]
\State  Randomly initialize $G_\theta$, $E_N$ and $E_C$.
\State  Pretrain $G_{\theta}$ using MLE on $\mathcal S$.
\State  Pretrain $E_N$ using negative samples from $G_{\theta}$ by minimizing the cross-entropy loss.
\State  Pretrain $E_C$ on $\mathcal L$ accordingly.
\For{number of training iterations}
  \For{G-steps}
    \State Sample $\hat Y$ from $G_{\theta}$
    \For{t in 1\ :\ T\ }
      \State Compute $Q(\hat y_t, s_{t-1})$ by Eq. (\ref{Equ_MC_R})
    \EndFor
    \State Update $G_\theta$ via Policy Gradient by Eq.~(\ref{Equ_gradient})
  \EndFor
  
  \For{teacherforce-steps}
    \State Update $G_{\theta}$ via MLE
  \EndFor
  
  \For{$E_N$-steps}
    \State Sample $\hat Y$ from new $G_{\theta}$ and sample $Y$ from $\mathcal{S}$
    \State Update $E_N$ via cross-entropy loss
  \EndFor
\EndFor
\State \Return $G_{\theta}$
\end{algorithmic}
\end{algorithm}

As forementioned, the $E_N$ needs adversarial training to get higher accuracy. Algorithm 1 shows the overall training process of the proposed approach, including the adversarial training of $E_N$. In Eq.~(\ref{Equ_gradient}), the ground-truth responses are not directly exposed to the generator in the training process. Practically, updating the generator $G_{\theta}$ only using the gradients from Eq.~(\ref{Equ_gradient}) leads to unstable training. The same issue is also reported in~\citeauthor{li2017adversarial} \shortcite{li2017adversarial}. To alleviate this issue, we follow \citeauthor{sutskever2014sequence} \shortcite{sutskever2014sequence} and use the Teacher Force strategy to train $G_{\theta}$, via MLE loss together with rewards from the evaluator.

\section{Experiments}
\label{sec_Exp}

\subsection{Datasets}

\subsubsection{Persona-Chat} We perform persona-based dialogue generation experiments on the Persona-Chat dataset~\cite{zhang2018personalizing}. The conversations are obtained from crowdworkers who were randomly paired and asked to act the part of a given persona. Also, the persona is created by another set of crowdworkers.
This dataset contains 164,356 utterances in 10,981 dialogues and has a set of 1,155 personas, each consisting of four or five persona texts. The testing set contains 1,000 dialogues (15,119 utterances) and 200 never seen before personas. We set aside 968 dialogues (15,705 utterances) together with its personas from the training set for validation. The final data has 10,000/968/1,000 dialogues for train/validate/test{\footnote{Note that the test set in ConvAI2 is different from the test set in \citeauthor{zhang2018personalizing} \shortcite{zhang2018personalizing} and is not publicly available.}}.

As reported in \citeauthor{zhang2018personalizing} \shortcite{zhang2018personalizing}, pretraining on larger datasets would yield better results. Thus we use another two million input-response pairs from OpenSubtitles to pretrain all models in our experiments, and we report this instead.


\subsubsection{DNLI} The recently released Dialogue Natural Language Inference dataset~\cite{WelleckDNLI} offers a new domain for NLI models. DNLI mainly consists of {\it utterance-persona} pairs, which are labeled as entailment ($\text{E}$), neutral ($\text{N}$), or contradiction ($\text{C}$). This dataset has 310,110/16,500/16,500 pairs for train/validate/test. Due to the length limit, we show other statistics of the DNLI dataset in the appendix.

\subsection{Baselines}
In the persona-based dialogue generation area, to the best of our knowledge, no previous work has explicitly modeled the consistency issue.
To evaluate our model, we compared the proposed approach with the following strong models:
\begin{itemize}
  \item{\bf S2SA} Seq2Seq is a generative dialogue model with the context attention mechanism~\cite{shang2015neural}. This is the only model {\bf without} persona information.

  \item{\bf Transformer} Transformer is one of the state-of-the-art sequence transduction models \cite{vaswani2017attention}. We concatenate persona texts to the message as its input.

  \item{\bf REGS} Reward for Every Generation Step is an adversarially trained model with Monte Carlo search for response generation~\cite{li2017adversarial}. We regard persona texts as dialogue context while training this model. 

  \item{\bf Per-S2S} This is a Seq2Seq model that prepends all persona texts to the input message~\cite{zhang2018personalizing}. 
    
  \item{\bf GPMN} Generative Profile Memory Network is a generative model that encodes persona as individual memory representations in a memory network~\cite{zhang2018personalizing}.
  
  \item{\bf DeepCopy} DeepCopy is a hierarchical pointer network, which extends the pointer-generator network to copy tokens from relevant persona texts~\cite{deepcopy}.
\end{itemize}

To make the following sections more concise, we abbreviate the proposed Reinforcement Learning based Consistent Dialogue Generation approach as {\bf RCDG}. Considering we have two different implementations ($E_{base}$ and $E_{bert}$) of the consistency module $E_C$, we use RCDG$_{base}$ and RCDG$_{bert}$ to denote implemented with $E_{base}$ and $E_{bert}$, respectively.

\subsection{Experimental Settings}
For the generator, both encoder and decoder are two-layer GRU with a hidden size 500. Embeddings of size 300 are randomly initialized and updated during training. Vocabulary size is 18,300, and other tokens are replaced with the {\it UNK} token. Encoder and decoder share the same vocabularies and embeddings. The model parameters are optimized using Adam with an initial learning rate of 0.0003. Learning rate decay is 0.98. Training minibatch size is 32. We set $\lambda$ to 0.4 and $N$ to 5. We implement the model in {\it OpenNMT-py}.

\subsection{Evaluation Metrics}

\subsubsection{Consistency Evaluation}
\label{sec_NLI_eval}
First, we evaluate the persona-consistency of different models.
\citeauthor{nli_eval} \shortcite{nli_eval} has shown that entailment techniques can be used as a surrogate for human judgment in evaluating dialogue consistency.
Following this work, we employ NLI model to automatically evaluate the persona-consistency of the generated responses. For a generated response $\hat Y$ and a set of persona texts $P=\{P_1,P_2,...,P_n\}$, an NLI model can assign an entailment category $l_i$ to each ($\hat Y$,$P_i$) pair, where $l_i \in \{\text{E},\text{N},\text{C}\}$. Then we simulate the human evaluator in deciding the entailment category between $\hat Y$ and $P$ by:
\begin{align}
\label{Equ_nli_eval}
NLI(\hat Y, P)=
&\left \{
\begin{aligned}
\text{E} \quad \quad if \ \ \ \ \ \ \ \text{E}\in L\\
\text{C} \quad \quad elif\ \ \ \, \text{C}\in L\\
\text{N} \quad \quad otherwise \ \ \
\end{aligned}
\right.
\end{align}
where $L=\{l_1,l_2,...,l_n\}$.

Considering we have used BERT as a consistency evaluator in the training process, it is not fair to use the same model again for evaluation. Thus we introduce another well-performed NLI model DIIN~\cite{gong2018DIIN}, as a third party, to evaluate all the dialogue models.

\subsubsection{Dialogue Quality Evaluation}
Second, the quality of generated dialogues is also an essential factor to consider.
We evaluate the dialogue quality of different models with the following metrics:
\begin{itemize}
\item{\bf Perplexity} Following \citeauthor{zhang2018personalizing} \shortcite{zhang2018personalizing}, we use perplexity (ppl) to measure the fluency of responses. Lower perplexity means better fluency.

\item{\bf Embedding metrics} Following \citeauthor{serban2016building} \shortcite{serban2016building}, we use Embedding Average (Ave.), Embedding Greedy (Grd.), and Embedding Extrema (Ext.) as evaluation metrics. These metrics are based on word embeddings, and they measure the relevance of a response regarding a target response. We use GoogleNews 300D word vectors.

\item{\bf Distinct} Following \citeauthor{li2015diversity} \shortcite{li2015diversity}, we calculate the token ratios of distinct 
bigrams (Distinct-2, abbreviated as Dst. for convenience). We use this metric to measure how diverse the responses are.
\end{itemize}


\begin{table}
\centering
\begin{tabular}{c|l|ll}
\toprule
\multicolumn{2}{c|}{\bf Model}&{\bf Dev}&{\bf Test}\\
\midrule
{\it Welleck}& InferSent & 85.82 & 85.68 \\
{\it et al. 2019}& ESIM & 86.31 & 88.20 \\
\midrule
{\it Models}& DIIN & 86.72 & 88.84 \\
{\it in}&$E_{base}$& 80.48 & 81.26 \\
{\it this work}&$E_{bert}$& {\bf 87.67} & {\bf 89.14} \\
\bottomrule
\end{tabular}
\caption{Accuracy of different models on the DNLI dataset.}
\label{tab:nli_acc}
\end{table}

\begin{table}[t]
\centering
\begin{tabular}{l|ll}
\toprule
{\bf Model}&{\bf Entail.(\%)}&{\bf Contr.(\%)}\\
\midrule
Human& 48.00 & 1.16$^*$  \\
\midrule
S2SA& 8.37 & 12.94  \\
GPMN& 12.98 & 11.53  \\
Per-S2S& 13.27 & 12.19  \\
REGS & 14.08 & 10.83  \\
Transformer & 14.20 & 9.00  \\
DeepCopy& 14.62 & 12.17  \\
\midrule

{RCDG$_{base}$} & 18.71 (28.0\%) & 5.93\ (34.1\%)\\
{RCDG$_{bert}$} & {\bf 19.07} (30.4\%) & {\bf 5.56} (38.2\%) \\
\bottomrule
\end{tabular}
\caption{NLI model-based persona-consistency evaluation results. $Entail.$ denotes entailment (the higher the better). $Contr.$ denotes contradiction (the lower the better). Best results are in bold, and the percentages in the parentheses are improvements regarding baselines' best results. $^*$ We show some contradiction examples of Human in the appendix.}

\label{tab:nli_eval_E_N_C}
\end{table}

\subsubsection{Human Evaluations}
\label{sec:human_eval}
In addition to the automatic evaluations, we also recruit five well-educated human judges to evaluate the generated responses.

Quantitatively evaluating the persona-consistency in generative models is a non-trivial task for humans. One major challenge is that the majority of the responses are neutral regarding the persona texts.
As we can see in the first row of Table \ref{tab:nli_eval_E_N_C}, even in the test set of Persona-Chat (from {\it Human}), half of the responses are neutral regarding the personas. This is plausible because many conversations in the real world do not ground on personas, such as greeting and question. With the limited sample size, we did not get statistically significant results in human evaluation when quantitatively evaluating the persona-consistency: the human judges labeled most of the sampled responses neutral.

Instead, we exploit human evaluations to verify the effectiveness of the model-based evaluation. Responses from all models are divided into three categories, and we randomly sample 150 response-persona pairs from each category. The judges are instructed to give a 5-scale score to each pair: {\bf 0}: definitely contradiction; {\bf 1}: potential contradiction; {\bf 2}: definitely neutral; {\bf 3}: potential entailment; {\bf 4}: definitely entailment. Note that the judges evaluate samples from each category (predicted by the DIIN), rather than from each model.

For dialogue quality, the evaluation is conducted following the usual practice. We sample 100 responses from each model and randomly shuffle them for judges. The five judges rate each response with a 3-scale criteria: {\bf0}: persona contradiction, irrelevant to the input, or grammatically broken; {\bf1}: the response reply to the message, but is not informative; {\bf2}: the response is relevant and informative.

\begin{figure}
    \centering
    \includegraphics[width=0.75\linewidth]{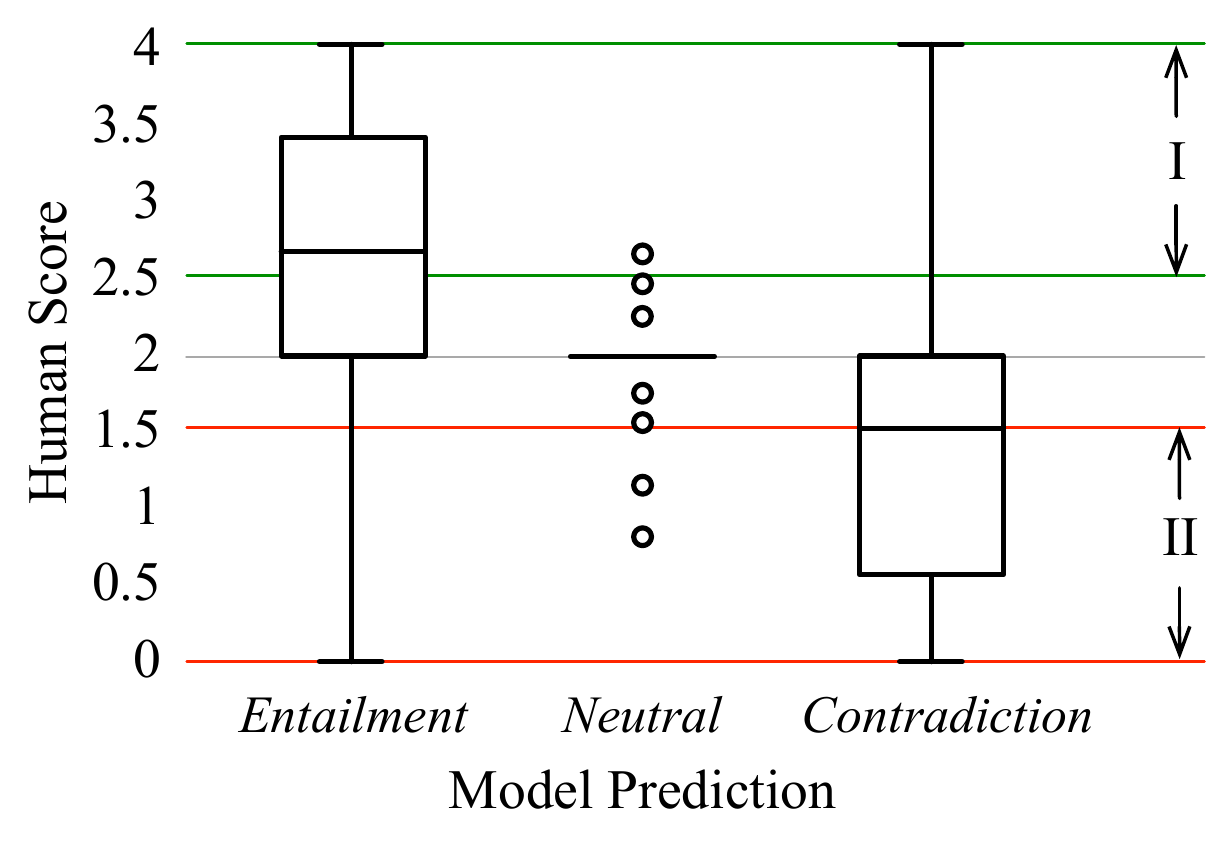}
    \caption{Boxplot of the human scores for consistency versus the model prediction categories. Three categories of model predictions are on the horizontal axis. With an average score greater or equal than 2.5, the area I is the score interval that is likely to be {\it Entailment}. Similarly, area II is likely to be {\it Contradiction}. This figure shows the correlation between human scoring and model prediction.}
    \label{fig:nli_box}
\end{figure}

\subsection{Results of Consistency}
Table~\ref{tab:nli_acc} shows the performance of different models on the DNLI dataset. The first two rows of results are reprinted from \citeauthor{WelleckDNLI} \shortcite{WelleckDNLI}. We implement the other three models. DIIN is the model for persona-consistency evaluation. The last two models ($E_{base}$ and $E_{bert}$) are two different implementations of the consistency module.

\subsubsection{Automatic Results} We report the model-based persona-consistency evaluation results in Table~\ref{tab:nli_eval_E_N_C}. 
With the explicit modeling of persona-consistency and reinforcement learning, our approach achieves the highest entailment percentage and a much lower contradiction percentage, compared with all other baselines.

The last two rows in Table \ref{tab:nli_eval_E_N_C} are the results of our approach, with different implementations of the consistency module. Both of them outperform other baselines significantly. Our RCDG$_{bert}$ gets better results, but this comes with higher computational costs, compared with our RCDG$_{base}$. The results could be interpreted to mean that the NLI signals work well, regardless of the NLI model structure.

\subsubsection{Human Validation} The human evaluation scores of each category are depicted in Figure \ref{fig:nli_box}. For the entailment category, more than half of the samples get an average score in the interval I, which means three judges agree that the sample is likely to be entailment or two judges agree and one of them is confident. For the neutral category, most samples get an average score of 2, and there are only a few outliers. This leads to the overlapping of the boxplot quartile lines. Figure \ref{tab:nli_eval_E_N_C} shows that the model-based evaluation is in a relatively good agreement with human evaluation. 
We have done a preliminary experiment in the evaluation of consistency, while a full study is beyond the scope of this paper.

\subsection{Results of Dialogue Quality}
We first report the automatic evaluation results of dialogue quality in Table \ref{tab:ppl_distinct_embedding}. Our methods are the best in the three embedding metrics, which indicates that our generated responses are most relevant to the ground truth. As our model is designed to address naturalness and consistency issues effectively, these results are within expectation. We notice that Transformer gets the best results in perplexity and distinct-2. It could be interpreted to mean that Transformer has a better language model compared with all other RNN based models. This also inspires us to use more advanced sequence models as our generator in future work. Except for the Transformer, our methods perform best in these RNN-based models. 

We report the human evaluation results in Table \ref{tab:human_repsonse_quality}. Our model has the highest ratio of 2, which means our generated responses are of higher quality. The Transformer also performs well in human evaluation, but it gets many 1 points. One reason could be that this model generates more questions rather than declarative sentences, which makes human judges feel less informative.


\begin{table}[t]
\centering
\begin{tabular}{l|lllll}
\toprule
{\bf Model}&{\bf ppl}&{\bf Ave.}&{\bf Grd.}&{\bf Ext.}&{\bf Dst.}\\
\midrule
DeepCopy         & 42.8 & 62.1 & 43.2 & 45.1 & 863 \\
Per-S2S          & 36.3   & 61.5 & 45.1 & 42.5 &   719\\
S2SA             & 34.8 & 59.8 & 41.9 & 43.5 & 473\\
GPMN             & 34.3 & 65.3 & 45.7 & 43.2 & 741 \\
REGS             & 33.6 & 64.3 & 44.2 & 44.8 &  1009\\
Transformer      & {\bf 28.1} & 63.4 & 43.9 & 43.6 & {\bf 1505}\\
\midrule
{RCDG$_{base}$}  & 30.2 & 66.7 & 46.9 & 46.4 & 1289\\
{RCDG$_{bert}$}  & 29.9 & {\bf 66.9} & {\bf 47.2} & {\bf 46.8} &  1275\\
\bottomrule
\end{tabular}
\caption{Automatic results, and Dst. is scaled by $10^{-4}$.}
\label{tab:ppl_distinct_embedding}
\end{table}

\begin{table}[t]
\centering
\begin{tabular}{l|lll|l|l}
\toprule
{\bf Model}&{\bf 0}&{\bf 1}&{\bf 2}&{\bf Avg}&{$\mathcal K$}\\
\midrule
S2SA        & 0.378 & 0.406 & 0.216 & 0.838 & 0.54 \\
GPMN        & 0.250 & 0.446 & 0.304 & 1.054 & 0.46 \\
Per-S2S     & 0.224 & 0.482 & 0.294 & 1.068 & 0.45 \\
REGS        & 0.242 & 0.440 & 0.318 & 1.076 & 0.42 \\
DeepCopy    & 0.224 & 0.450 & 0.326 & 1.102 & 0.48 \\
Transformer & 0.212 & 0.458 & 0.330 & 1.118 & 0.43 \\
\midrule
{RCDG$_{base}$}  & 0.182 & 0.440 & 0.378 & 1.196 & 0.50 \\
{RCDG$_{bert}$} & 0.180 & 0.436 & 0.384 & 1.204 & 0.47 \\
\bottomrule
\end{tabular}
\caption{The results of human evaluation for response quality, together with the Fleiss Kappa ($\mathcal K$). The $\mathcal K$ coefficient between 0.41 and 0.6 means moderate agreement.}
\label{tab:human_repsonse_quality}
\end{table}

\subsection{Ablation Study}
As the proposed model achieves better performance than previous approaches, we conduct an analysis to gain further insight on how the integration of different modules helps the response generation. We report the results in Table \ref{tab:ablation}.

As we can see, the performance of the vanilla generator is not outstanding. With the help of the adversarially trained naturalness module $E_N$, the dialogue quality is improved. Meanwhile, if we directly apply reinforcement learning without the naturalness module, although the consistency of the generated response is improved, the quality has decreased significantly, as shown in the $+ E_{base}$ and $+ E_{bert}$. 
When we integrate the naturalness module and the consistency module, the performance achieves the best.

Finally, we show some generated examples in Table \ref{tab:cases}.

\section{Conclusion and Future Work}

In this paper, we consider modeling the persona-consistency in open-domain dialogue generation by exploiting natural language inference. To this end, we cast the task as a reinforcement learning problem and leverage natural language inference signals in the deep generative model. We demonstrate the effectiveness of our approach in comparison with several baselines by experiments on the Persona-Chat dataset. In the future, we plan to apply our model to larger scale datasets. Furthermore, we plan to use more advanced generators in our approach to achieve higher performance.

\begin{table}[t]
\centering
\begin{tabular}{l|lllll}
\toprule
{\bf Model}&{\bf ppl}&{\bf Ave.}&{\bf Dst.}&{\bf E.(\%)}&{\bf C.(\%)}\\
\midrule
{Generator}  & 34.5 & 63.8 & 823 & 13.23 & 12.08 \\
\midrule
{+ $E_{N}$}     & 32.3 & 64.8 & 1078 & 14.01 & 11.32 \\
{+ $E_{base}$}  & 40.9 & 60.3 & 639 & 16.98 & 7.52 \\
{+ $E_{bert}$}  & 41.3 & 59.9 & 517 & 17.13 & 7.19 \\
{+ $E_{N}$ + $E_{base}$}  & 30.2 & 66.7 & {\bf1289} & 18.71 & 5.93 \\
{+ $E_{N}$ + $E_{bert}$}  & {\bf 29.9} & {\bf66.9} & 1275 & {\bf19.07} & {\bf5.56} \\
\bottomrule
\end{tabular}
\caption{Ablation test of different modules.}
\label{tab:ablation}
\end{table}

\begin{table}[t]
\centering
\begin{tabular}{r|l}
\toprule
{\bf Persona}&{I'm a {\bf 12} years old girl}\\
{\bf Input}&{Hi there! I am age 24, how about you?}\\
\midrule
Transformer & {\bf 30}. I want to be a baseball announcer.\\
REGS        & I'm {\bf 30} years old. \\
GPMN        & I am {\bf 32} and older than you.\\
DeepCopy    & I'm {\bf 12} years old.    \\
Ours        & I am {\bf 12} years old. \\
\midrule
{\bf Persona}&{I only listen to {\bf country music}}\\
{\bf Input}&{What sort of music do you listen to?}\\
\midrule
Transformer & What kind of music do you like? \\
REGS        & I love {\bf country}. Especially tyler farr swoon.  \\
GPMN        & I love {\bf rap}, Tupac is my favorite. You?  \\
DeepCopy    & My sort of music pays well. \\
Ours        & My favorite is {\bf country}, what about you? \\
\bottomrule
\end{tabular}
\caption{Sampled dialogues from different models.}
\label{tab:cases}
\end{table}

\section{Acknowledgments}
The paper is supported by the National Natural Science Foundation of China under Grant No.61772153. In addition, we want to acknowledge the Heilongjiang Province Art Planning Project 2019C027 and the Heilongjiang Province Social Science Research Project 18TQB100. We also want to thank all the anonymous reviewers for their comments.

\bibliographystyle{aaai}
\bibliography{AAAI-SongHY.1681}

\end{document}